\documentclass[lettersize,journal]{IEEEtran}
\usepackage{amsmath,amsfonts}
\usepackage{algorithmic}
\usepackage{algorithm}
\usepackage{booktabs}
\usepackage{multirow}
\usepackage{multicol}
\usepackage{array}
\usepackage{textcomp}
\usepackage{stfloats}
\usepackage{url}
\usepackage{verbatim}
\usepackage{graphicx}
\usepackage{cite}
\usepackage[backref]{hyperref} 
\usepackage{makecell}
\usepackage{array}
\begin{document}

\title{A Data Synthesis Method Driven by Large Language Models for Proactive Mining of Implicit User Intentions in Tourism}

\author{Jinqiang Wang, 
Huansheng Ning, ~\IEEEmembership{Senior Member,~IEEE,}
Tao Zhu,
and Jianguo Ding,  ~\IEEEmembership{Senior Member,~IEEE}
\thanks{Jinqiang Wang and Huansheng Ning are with the School of Computer \& Communication Engineering, University of Science and Technology Beijing, 100083 China. e-mail: jqwang@xs.ustb.edu.cn, ninghuansheng@ustb.edu.cn. \emph{(Corresponding author: Huansheng Ning)}}
\thanks{Tao Zhu is with the School of Computer Science, University of South China, 421001 China. e-mail: tzhu@usc.edu.cn.}
\thanks{Jianguo Ding is with  with the Department of Computer Science, Blekinge institute of Technology, Karlskrona, Sweden. e-mail: jianguo.ding@bth.se.}
\thanks{The code and dataset can be found at: \url{https://github.com/jqwangai/SynPT}}}



\maketitle

\begin{abstract}
In the tourism domain, Large Language Models (LLMs) often struggle to mine implicit user intentions from tourists’ ambiguous inquiries and lack the capacity to proactively guide users toward clarifying their needs.
A critical bottleneck is the scarcity of high-quality training datasets that facilitate proactive questioning and implicit intention mining. While recent advances leverage LLM-driven data synthesis to generate such datasets and transfer specialized knowledge to downstream models, existing approaches suffer from several shortcomings: (1) lack of adaptation to the tourism domain, (2) skewed distributions of detail levels in initial inquiries, (3) contextual redundancy in the implicit intention mining module, and (4) lack of explicit thinking about tourists' emotions and intention values.
Therefore, we propose SynPT (A Data Synthesis Method Driven by LLMs for Proactive Mining of Implicit User Intentions in the Tourism), which constructs an LLM-driven user agent and assistant agent to simulate dialogues based on seed data collected from Chinese tourism websites. This approach addresses the aforementioned limitations and generates SynPT-Dialog, a training dataset containing explicit reasoning. The dataset is utilized to fine-tune a general LLM, enabling it to proactively mine implicit user intentions.
Experimental evaluations, conducted from both human and LLM perspectives, demonstrate the superiority of SynPT compared to existing methods. Furthermore, we analyze key hyperparameters and present case studies to illustrate the practical applicability of our method, including discussions on its adaptability to English-language scenarios. All code and data are publicly available.
\end{abstract}


\section{Introduction}
\IEEEPARstart{L}{arge} language models (LLMs) have demonstrated significant capabilities across various domains, including healthcare \cite{10840322, wang2024survey}, education \cite{parker2024large}, code generation \cite{10658990}, and tourism \cite{wei2024tourllm}. Particularly, LLMs can perform tasks such as intelligent customer service and attraction explanations in the context of tourism. However, most existing LLM applications passively respond to user instructions rather than engaging in proactive interaction. When faced with ambiguous queries, they are generally unable to proactively seek clarification or mine users' implicit intentions \cite{manggala2023aligning}. For instance, when a user requests, “Recommend me some Western restaurants”, the explicit intention (type of restaurant) is clear, but the implicit intention, such as budget and preferred location, remains unstated. Consequently, LLMs often struggle to fulfill user expectations when executing inherently ambiguous instructions.
\par
\begin{figure*}
    \centering
    \includegraphics[width=0.8\linewidth]{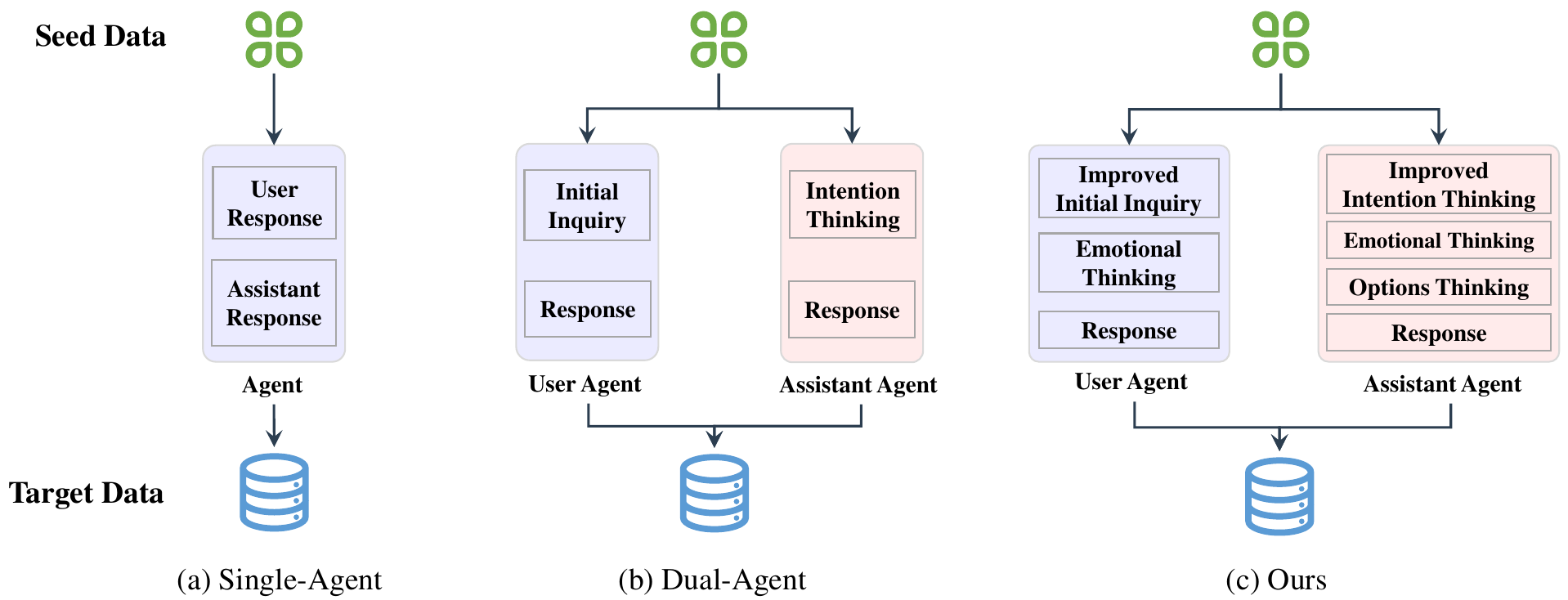}
    \caption{Various Methods of Data Synthesis. (a) This method utilizes a single agent to synthesize data. a representative work is \cite{xu2023baize}. (b) This method employs two interacting agents for data synthesis. a representative work is \cite{qian2024tell}. (c) In comparison to (b), this approach introduces emotion thinking and option thinking modules, and further optimizes the mechanisms of the initial inquiry and intention thinking modules. In (a) and (b), each agent is played by a single LLM. In (c), each module is played by a single LLM, and each agent consists of multiple such modules. }
    \label{fig:introduction}
\end{figure*}

Recent research \cite{deng2025proactive, zhang2024clamber} investigates approaches to improving the proactive questioning capabilities of LLMs, enabling them to clarify ambiguous user instructions and mine implicit intentions. However, existing approaches \cite{zhang2024ask, deng2023survey} often rely on complex workflows involving multiple LLM reasoning steps, resulting in significant computational overhead and resource inefficiency. The LLM-driven data synthesis approach presents a feasible solution \cite{xu2023baize, lu2025proactive}. As illustrated in Fig. \ref{fig:introduction}(a), this method first employs an LLM to generate a training set with the capability of proactively mining users implicit intentions. Subsequently, the synthesized training set is used to train the target model, equipping it with the same capability. During downstream tasks, the model requires only a single inference to proactively mine implicit user intentions and proactively ask users. We refer to this paradigm as the Single-Agent. In addition, Mistral-Interact \cite{qian2024tell} employed an LLM-driven user agent and assistant agent to simulate dialogues, generating a training set capable of proactively mining implicit intentions, as illustrated in Fig. \ref{fig:introduction}(b). In this method, the assistant agent mines implicit intentions within user instructions and proactively queries the user agent for clarification. The user agent then responds based on contextual information. This method is termed the Dual-Agent paradigm. \textbf{However, existing data synthesis methods present several issues:} (1) Prior studies does not involve the tourism scenario and lacks the capability to proactively mine implicit user intentions in this scenario. (2) The distribution of detail levels in initial inquiries is overly concentrated, restricting diversity in expression granularity. (3) In multi-round dialogue scenarios, the assistant agent repeatedly loads historical dialogue when mining implicit user intentions, causing a linear increase in input sequence length and higher computational overhead. (4) The assistant agent lacks the explicit thinking for user emotions and potential intention value options. 
\par
To address the above issues, we propose \textbf{SynPT}, an LLM-driven data \textbf{syn}thesis method for \textbf{p}roactive mining of implicit user intentions in \textbf{t}ourism, which is schematically shown in Fig. \ref{fig:introduction}(c). \textbf{Specifically, we present the following key innovations: }(1) Focusing on the Chinese tourism domain, SynPT is the first to leverage tags from tourism websites as seed data for synthesizing a multi-round conversation dataset, thereby facilitating proactive implicit intention mining in this domain. Compatibility with English scenarios will be discussed in the case study section. (2) In designing the user agent, we implement a probabilistic control mechanism to ensure that the granularity of initial inquiries aligns with real-world distributions. (3) To enhance the assistant agent, we propose a memory stack mechanism within the intention thinking module. This mechanism stores the current intention and replaces historical dialogue records with memory stack content, thereby reducing contextual redundancy in subsequent intention thinking. (4) Additionally, the assistant agent incorporates user emotion thinking and intention value thinking modules to better comprehend users' implicit intentions. Using the SynPT-Dialog dataset generated by SynPT, we fine-tuned the Qwen2.5-7B-Instruct \cite{yang2024qwen2} model to develop Qwen-PT. This model proactively mines users implicit intentions in Chinese travel scenarios without explicit prompt.
\par
We evaluate the capability of Qwen-PT to proactively mining implicit user intention, thereby demonstrating the effectiveness of SynPT. To ensure generalization, we constructed a test set comprising samples from cities distinct from those in the training set. We then conducted a comparative analysis of different data synthesis methods using both LLM-as-a-judge \cite{zheng2023judging} and human evaluation. The experimental results demonstrate that the Qwen-PT model, trained using data generated by SynPT, outperforms models trained on data generated by other methods (Single-Agent and Dual-Agent) in proactively identifying implicit intentions within Chinese travel scenarios. Furthermore, it surpasses LLMs accessed through commercial APIs, including DeepSeek-R1 \cite{guo2025deepseek} and Doubao-Pro, across most metrics. We also investigate the impact of different key hyperparameters, backbone networks, and the choice of LLMs for both user and assistant agents on overall model performance. Finally, we present case studies demonstrating practical applications of Qwen-PT, with a focus on its efficacy in mitigating negative emotions elicited by rhetorical questions, enhancing task scalability, and exploring adaptability in English-dominated environments.
\par
The contributions of this paper are as follows.
\par
\begin{enumerate}
\item \textbf{A novel idea for data synthesis.} This study is the first to explore the feasibility of transforming tags from Chinese travel websites into multi-round dialogue data capable of proactivity mining implicit user intentions.
\par
\item \textbf{Advanced data synthesis methodology.} We propose an enhanced LLM-driven data synthesis method for proactively mining implicit user intention. By integrating probabilistic control, intention memory stacks, user emotion thinking, and intention value thinking, the proposed approach significantly outperforms existing data synthesis methods (Single-Agent and Dual-Agent) in proactivity mining implicit user intention.
\par
\item \textbf{Comprehensive Resource Release.} We open-source the code for the SynPT synthesis method and developed an easy-to-use data synthesis tool. Additionally, we release the SynPT-Dialog synthetic dataset and the Qwen-PT model weights.
\end{enumerate}
\par
The remainder of this paper is structured as follows. Section II provides an overview of related research work. Section III describes in detail our proposed data synthesis method, SynPT. Section IV describes the specific setup of the experiments. Section V analyzes and discusses the main experimental results. Section VI provides a systematic discussion of the hyperparameters of SynPT. Section VII demonstrates the practical application of Qwen-PT through case studies. Finally, Section VIII summarizes the study and outlines directions for future research.

\section{Related work}
\subsection{Data synthesis methods driven by LLMs}
The success of LLMs heavily depends on the availability of large-scale, high-quality data, and this requirement is equally critical for domain-specific LLMs \cite{gunasekar2023textbooks, NEURIPS2024_ed165f2f}. LLMs can extract complex patterns from large datasets, generate synthetic data that closely matches real-world distributions, and incorporate custom-defined rules. Furthermore, data synthesis methods driven by LLMs can be employed for model distillation by generating synthetic data using powerful generalized models to improve the performance of smaller or specialized models \cite{wang2024survey-synthesis, NEURIPS2024_0356216f}. The Phi-1 model family \cite{li2023textbooks} has shown that small quantities of high quality textbook and exercise data generated by GPT-3.5 are sufficient to train highly capable models.
\par
Several LLM-driven data synthesis methods have been proposed. Self-Instruct \cite{wang2023self} prompts an LLM to generate new data based on a manually curated pool of seed tasks. The generated data is then filtered and incorporated back into the pool of seed tasks. This iterative process is repeated multiple times to produce a substantial volume of data. Baize \cite{xu2023baize} utilizes Q\&A pair data as a seed dataset to prompt an LLM to generate multi-turn conversations between humans and AI assistants. These generated conversations are subsequently used to fine-tune an LLM, thereby improving its conversational and problem 
solving abilities. Baize employs two data synthesis strategies: (1) generating an entire multi-turn conversation in a single call to ChatGPT, and (2) generating one turn of dialogue per call for a single role (either human or assistant), then sequentially combining these into a complete multi-turn conversation. Similarly, MAGPIE \cite{xu2025magpie} generates conversations by sequentially calling an LLM for each role, ultimately constructing a complete multi-turn dialogue. In contrast, UltraChat \cite{ding2023enhancing} employs two independent instances of ChatGPT to generate informative and realistic multi-turn conversations. One instance acts as the user by producing queries, while the other simulates the AI assistant by generating responses. PlatoLM \cite{kong2024platolm} trains a user simulator on the ShareGPT dataset, which is then used to generate a multi-turn conversation dataset by enabling natural interactions between the simulator and ChatGPT. Similarly, Abbasiantaeb et al. \cite{abbasiantaeb2024let} investigates the use of LLMs to simulate conversational question answering between humans. Specifically, it employs GPT-4 \cite{achiam2023gpt} to simulate a student-teacher dialogue, where the student poses questions based on a Wikipedia article, and the teacher generates and verifies the accuracy of the responses. This study \cite{ge2024scaling} adopts a role based approach to generate diverse synthetic data. It first extracts diverse roles from large scale web data and then guides LLM to synthesize data from the perspective of the roles. LawGPT \cite{zhou2024lawgpt} produces high quality legal reasoning data through strategies such as leveraging a knowledge base, implementing error correction, and conducting validation.
\par
However, existing data synthesis methods fail to incorporate proactive user inquiry or account for critical scenarios such as implicit user intention mining, emotional shifts, and intention value suggestions. To address these limitations, we propose a novel data synthesis method that leverages publicly available travel websites as seed data. This approach generates multi-turn dialogues with proactive questioning capabilities and integrates reasoning trajectories, including user intention thinking, user emotion thinking, and intention value suggestions.

\subsection{Proactive user intention mining}
Currently, most LLM based agents require explicit human instructions to initiate and execute tasks, remaining inactive until further input is provided. This paradigm limits their capacity to proactively assist or autonomously provide services without human instructions \cite{deng2023survey}. Recent research has focused on enhancing the proactivity of agents. Deng et al. \cite{deng2023prompting, zhang2024clamber} employed an LLM to generate user-oriented inquiry questions aimed at clarifying vague user intentions. Similarly, Naszad et al. \cite{manggala2023aligning} examined the boundaries between vague and clear intentions based on real conversations, exploring when agents should proactively pose clarifying questions to users. Zhang et al. \cite{zhang2024ask} proposed a proactive agent for planning tasks to better understand the user's intention by proactively seeking information to refine the user's task. However, the agent lacks the chain-of-thought reasoning process in responses, limiting its generalizability to other tasks. Mistral-Interact \cite{qian2024tell} leveraged the interaction between user agents and assistant agents to generate training samples for proactive questioning, thereby enhancing the model's ability to mine implicit user intentions. Lu et al. \cite{lu2025proactive} developed more complex environments to generate samples that support proactive prediction of user intentions in subsequent tasks, thereby enhancing the model's capability for proactive questioning. The MDD-5K \cite{yin2025mdd} framework facilitates interactions between a doctor agent and a patient agent, simulating real-world clinical dialogues involving mental health disorders. In these interactions, the doctor agent guides the conversation by employing a dynamic diagnosis tree to structure its inquiries, while the patient agent generates responses based on authentic patient cases and synthetic experiences. The resulting dialogue data improves the model's capacity for proactive questioning and diagnostic reasoning in mental health scenarios.
\par
However, the existing work does not deeply explore Chinese travel scenarios and does not think about user emotions and reference options of intention. To bridge these gaps, we propose a data synthesis method that generates data containing the above thought process. The resulting data enables a model to think about user implicit intention, emotions, and reference options of intention values before proactively initiating inquiries.

\section{Methods}

\subsection{Overview}
\begin{figure*}
    \centering
    \includegraphics[width=0.9\linewidth]{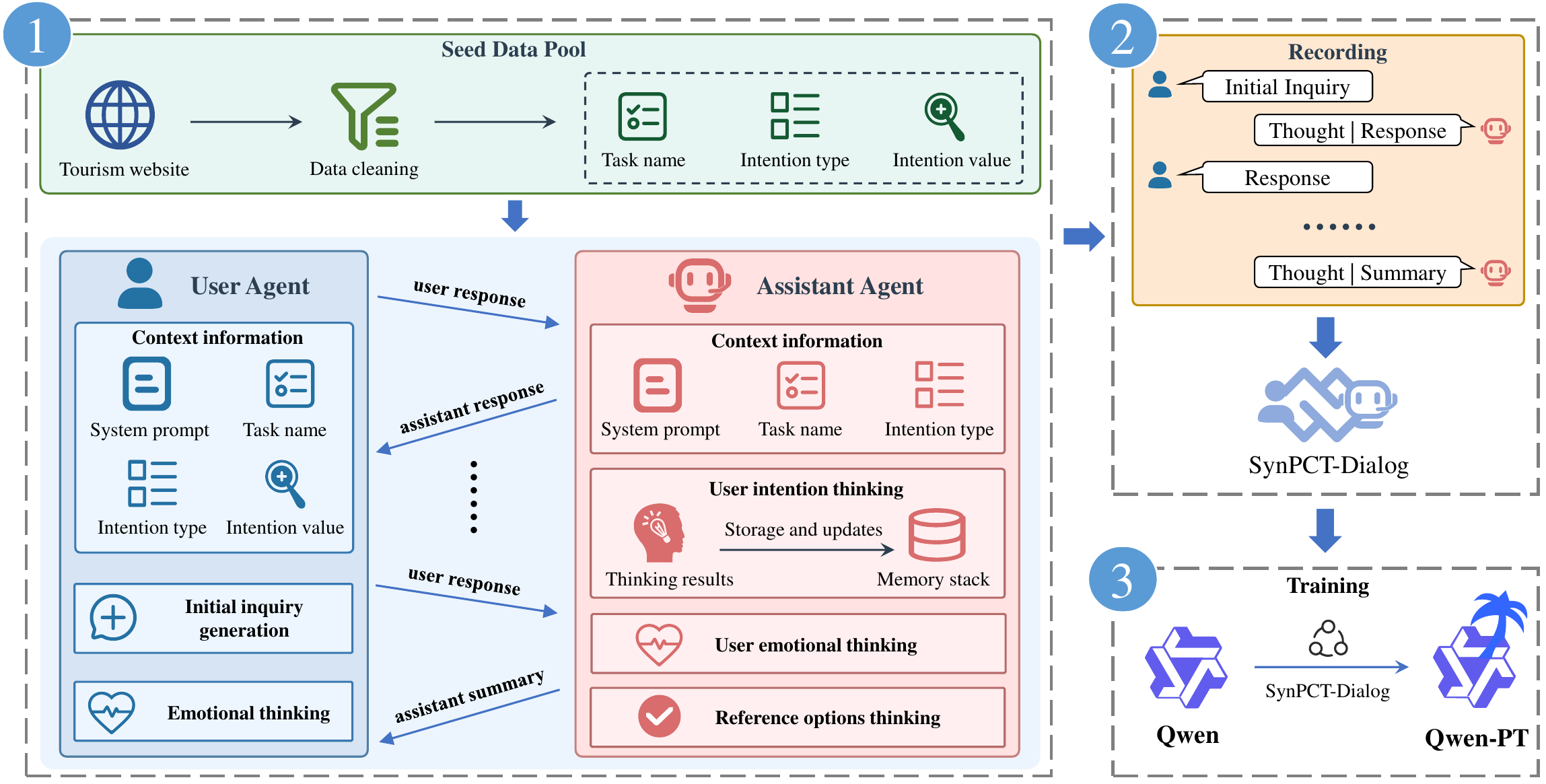}
    \caption{Overview of SynPT Methodology and Downstream Processing. It comprises four components: the seed data pool, the user agent, the assistant agent, and the recording component. The recording component collects dialogue data generated by interactions between the user agent and the assistant agent, forming a multi-turn dialogue dataset capable of proactively mine implicit user intention. The Qwen model is subsequently fine-tuned on this dataset, resulting in Qwen-PT.}
    \label{fig:method}
\end{figure*}

As illustrated in Fig. \ref{fig:method}, we propose a data synthesis method named SynPT, which comprises four core components: the \textbf{seed data pool}, the \textbf{user agent}, the \textbf{assistant agent}, and the \textbf{recording} component. The seed data pool is responsible for collecting and constructing the contextual environment for dialogues between the user agent and the assistant agent, including the tour task, corresponding intention types, and intention values. The user agent plays the role of the user communicating with the system assistant, while the assistant agent plays the role of the system assistant proactively asking the user for implicit intentions. The recording component captures the interaction process between the user and the assistant agent, including their reasoning content, to generate a multi-turn conversational dataset with the feature of proactively mining implicit user intentions. Compared to prior work \cite{qian2024tell}, our approach introduces several key improvements, particularly in the seed data pool (an innovation in application scenarios) and the modules for initial inquiry generation, user intention thinking, emotional thinking, and reference option thinking. The following subsections provide a detailed explanation of these enhancements.

\par
The process of this data synthesis method is outlined as follows. \textbf{First}, a data instance is sampled from the seed data pool, which contains the task name $I^n$, intention type $I^t$, and intention value $I^v$. \textbf{Next}, the user agent $Agent^U$ generates an initial inquiry based on the sampled data and sends it to the assistant agent. This can be represented as:
\begin{equation}
inquiry = Agent^U(I^n,I^t,I^v)
\end{equation}
\par
Based on the initial inquiry, the task name $I^n$, and the intention type $I^t$, the assistant agent performs three types of reasoning: user intention reasoning (identifying the task name the user is inquiring about, recognizing explicit user intentions, and mining any implicit intentions), user emotion reasoning (assessing whether the user is willing to continue the conversation), and reference option reasoning (estimating the most probable intention value corresponding to the current inquiry type). Based on these reasoning, the assistant agent $Agent^A$ generates a response $R_{1}^{A}$ and sends it to the user agent. That is: 
\begin{equation}
T_{1}^{A}, R_{1}^{A} = Agent^A(inquiry, I^n, I^t)
\end{equation}
Let $T_{1}^{A}$ denote the assistant agent's first-round reasoning process, and $R_{1}^{A}$ denote the assistant agent's first-round response.
\par
Upon receiving the response from the assistant agent, the user agent initiates an emotion thinking process. If the user agent detects that the assistant agent has repeatedly inquired about implicit user intentions within the dialogue history $H$, it may express unwillingness to continue the interaction in its response. The formal representation of the process is:
\begin{equation}
R_{2}^{U} = Agent^U(I^n,I^t,I^v, H)
\end{equation}
Where $R_{2}^{U}$ denotes the response of the user agent in the second round of dialog.
\par
Finally, the communication between the two agents terminates when either the termination condition is met or the maximum number of communication rounds is reached. The termination condition is triggered if (1) the user intention thinking indicates no remaining unprovided implicit intentions or (2) the user emotion thinking suggests reluctance to continue the interaction. Additionally, this study focuses on enhancing the performance of proactively mining implicit user intentions rather than improving the resolution of user inquiries. \textbf{Consequently, the assistant agent's final response is a summary of the user's intention, rather than a direct response to user inquiries.}
During the communication process, the recording component captures both the user agent’s responses and the assistant agent’s reasoning processes and replies, thereby constructing a multi-turn dialogue dataset. For a given data sample with k dialogue turns, the example $X$ can be represented as:
\begin{equation}
X = \{inquiry, [T_{1}^{A}, R_{1}^{A}], R_{2}^{U}, ... , R_{k}^{U}, [T_{k}^{A}, R_{k}^{A}]\}
\end{equation}

\subsection{Seed data pool}
We begin by collecting seed data from Chinese travel websites such as Ctrip and Qunar, with the goal of stimulating interactions between a user agent and an assistant agent around the implicit user intention. To identify relevant travel-related tasks, we apply the following three criteria: (1) the task is one that users might seek assistance with from a LLM; (2) the task pertains to one of the six key aspects of travel——food, accommodation, transportation, sightseeing, shopping, or entertainment; and (3) the task is associated with a rich set of tags on the target websites. These tags are typically designed to encourage user engagement, such as filtering content in travel blogs through common tags like ``trip duration" and ``average budget". In this study, we treat these tags as proxies for user intention and collect them accordingly. To the best of our knowledge, this is the first study to utilize these tags as seed data for data synthesis.
\par
As shown in Table 1, we identified six tasks to construct the seed data pool and defined its structural components: task name, intention type, intention value, and reference answer. Notably, reference answers are utilized only in the case study presented in Section \ref{sec:executing user intentions}, as the primary objective of this study is to proactively mine implicit user intentions rather than to directly address their problems. This study employs a web crawler to collect seed data of eight representative tourist cities in China: Beijing, Shanghai, Guangzhou, Chengdu, Xi'an, Hangzhou, Nanning, and Guilin. After removing outliers, a substantial volume of seed data was obtained. Due to token consumption constraints, we conducted random sampling to select a manageable subset. Finally, the number of seed data involved in this paper is shown in Table \ref{tab:task}.

\begin{table*}[]
\caption{Details of seed data. Detailed explanations of the intention types and seed datasets can be found in our published repository.} \label{tab:task}
\begin{tabular}{@{}l>{\raggedright\arraybackslash}p{0.66\linewidth}>{\centering\arraybackslash}p{0.1\linewidth}@{}}
\toprule
Task Name       & Intention Type (Tags)                                                          & Count \\ \midrule
Travel Planning & Destination, Duration, Time, Per Capita Budget, Companions, Travel Style & 290        \\ \midrule
Attraction  Recommendation &
  Destination, Location, Attraction Type &
  200 \\ \midrule
Restaurant Recommendation &
  Destination, Location, Cuisine Type, Budget Level &
  250 \\ \midrule
Accommodation Booking &
  Destination, Location, Check-in/Check-out Dates, Hotel Star Rating, Accommodation Type, Room Type &
  254 \\ \midrule
Shopping Venue Inquiry &
  Destination, Location, Shopping Type &
  241 \\ \midrule
Train Ticket Reservation &
  Departure, Destination, Departure Date, Departure Time Window, Train Type, Seat Type &
  291 \\ \bottomrule
\end{tabular}
\end{table*}

\subsection{Initial inquiry}
The initial inquiry in prior research was directly generated by a LLM, expressed as:
\begin{equation}
    query=f(P_q,I^n,I^t,I^v)
\end{equation}
where $P_q$ denotes the prompt for generating the initial inquiry, while $I^n,I^t,I^v$ represent the task name, intention type, and intention value, respectively. 
\par
However, this prompt-based approach limits the flexibility to control the level of detail in the initial inquiry, specifically regarding the number of included intention types.
To this end, we conducted preliminary experiments in which a LLM was instructed to generate an initial inquiry containing a specified number of intentions. The generated outputs were then evaluated to assess whether they aligned with the predefined number of intentions. The experimental results are shown in Table \ref{tab:inquiry}, and input-output examples are available in our GitHub repository. The results show that models without reasoning chains largely failed to complete the task. Among the models with reasoning chains, o3-mini performed relatively well, but it still made two errors and exhibited high invocation costs and long reasoning times. Furthermore, our analysis of the incorrect initial inquiries revealed that they consistently included more intentions than the predefined number.
Therefore, this prompt-based approach biases the generated initial inquiries toward those containing a greater number of intentions, thereby neglecting inquiries with fewer intentions. As a result, the LLM trained on such synthetic data exhibit reduced generalization performance.
 
{
\setlength{\tabcolsep}{13pt}
\begin{table*}[]
\caption{Evaluation of Whether Initial Inquiries Generated by LLMs Cover a Predefined Number of Intentions Using 10 Prompt Templates}
\label{tab:inquiry}
\begin{tabular}{@{}lccccccc@{}}
\toprule
\multirow{2}{*}{Models} & \multicolumn{4}{c}{w/o Reasoning}     & \multicolumn{3}{c}{w/ Reasoning}   \\ \cmidrule(l){2-5} \cmidrule(l){6-8} 
                  & Doubao-Pro & DeepSeek-V3 & Qwen-max & GPT-4o & DeepSeek-R1 & o3-mini & Claude-3-7-Sonnet-Thinking \\ \midrule
Score             & 1/10       & 1/10        & 0/10     & 0/10              & 1/10        & 8/10    & 7/10                       \\ \bottomrule
\end{tabular}
\end{table*}
}

\par
Based on the identified problem, we propose an improvement that involves selectively providing a subset of the intention type $I^{t}_{inedx}$  and intention value $I^{v}_{inedx}$ using probabilistic control. In contrast to previous approaches, the full intention type $I^t$  and intention value $I^v$ are no longer supplied to the LLM. The probabilistic control mechanism is described in detail below.
\par
The first step involves controlling the number of intentions in the initial inquiry. Let the candidate set of intention counts be $\mathcal{N} = {0, 1, 2, \dots, N}$, where $N$ denotes the maximum number of intentions for a given task. For each $n \in \mathcal{N}$, the probability density function of the normal distribution is calculated as follows:
\begin{equation}
    \phi(n; \mu, \sigma) = \frac{1}{\sqrt{2\pi \sigma^2}} \exp\left( -\frac{(n - \mu)^2}{2\sigma^2} \right)
\end{equation}
Here, the distribution mean \( \mu \) and standard deviation \( \sigma \) are hyperparameters that govern the expected number of intentions in the initial inquiry. To obtain a valid discrete probability distribution over the set \( \mathcal{N} \), the normal distribution is normalized as follows:
\begin{equation}
    P(n) = \frac{\phi(n; \mu, \sigma)}{\sum_{m=0}^{N} \phi(m; \mu, \sigma)}
\end{equation}
\par
Next, an intentional quantity $m$ is drawn as a discrete random variable from the set $\mathcal{N}$ according to a probability distribution $P(n)$, expressed as:
\begin{equation}
    m \sim P(n) \quad n \in \mathcal{N}
\end{equation}

\par
Subsequently, non-playback sampling is employed to select $m$ distinct intentions from the intention type set $I^{t} = \{t_1, \dots, t_N \}$ and the intention value set $I^v = \{v_1, \dots, v_N\}$. In this process, each candidate's intention is assigned a user-defined raw probability $k_i$ reflecting its likelihood of being included in the initial inquiry. To ensure normalization, the probabilities are adjusted using the following equation:
\begin{equation}
    q_i = \frac{k_i}{\sum_{j=1}^N k_j}, \quad i = 1, 2, \dots, N
\end{equation}
Randomly select $m$ indices from the set $\{1,2,\dots,N\}$ without replacement, following the probability distribution $\{q_1, q_2, \allowbreak \dots, q_N\}$. This sampling process is denoted as:
\begin{equation}
    index= Sample(\{1,2,…,N\},m,\{q_1,q_2,…,q_N\})
\end{equation}
The set of intention types $I_{t}^{inedx}$ and the set of intention values $I_{v}^{inedx}$,  sampled by index, can be obtained. Subsequently, the improved initial query can be derived as:
\begin{equation}
    \text{query}' = f(P_q,I^n,I^t_{\text{index}}, I^v_{\text{index}})
\end{equation}
\par
This strategy enhances the generation probability of initial inquiries that include either zero or all  intentions by adjusting the distribution's mean and standard deviation. Additionally, it ensures that most generated inquiries align with human preferences by customizing the intention selection probability.

\subsection{User intention thinking}
The user intentions thinking module generates two key elements: the user intention and the implicit intention type that requires subsequent clarification from the user. The user intention comprises three aspects: (1) the task under consultation, (2) the intention types and their corresponding values that have already been provided, and (3) the intention types yet to be specified—termed implicit intentions. The previous approach employs the complete conversation history $H$ as input and extracts the user’s intention using a LLM $f$, formally expressed as:
\begin{equation}
    I = f(P_I,H)
\end{equation}
Here, $P_I$ represents the system prompt for the module. In multi-round dialogue scenarios, the model must continuously recognize the user's intention and update the recognition results in each round. However, supplying the complete dialogue history in every round introduces significant redundancy, increasing token consumption. Moreover, excessive historical context can reduce the LLM's accuracy in intention extraction. To address this, we propose a memory stack $S$ to store user intentions, enabling the model to focus on the current round of intention mining while reducing contextual redundancy and improving correctness. The stack is initialized as follows:
\begin{equation}
    S_1 = f(P_{init}, H_1^U)
\end{equation}
Where $P_{init}$ is the system prompt for initializing the memory stack and $H_1^U$ denotes the first round of user consultation. For subsequent rounds, the update strategy is defined as:
\begin{equation}
    {S_i, T_i} = f(P_u, H_i^U, H_{i-1}^A,  S_{i-1})
\end{equation}
$H_{i-1}^A$ represents the assistant agent's response in the \text{(i-1)-th} round of dialogue, while $H_i^U$ denotes the user agent's response in the \text{i-th} round. $P_u$ indicates the prompt for this strategy.
$T_i$ represents the intention type that the assistant inquires about from the user in the i-th round. 
The pair $\{S_i, T_i\}$ serves as the output of this module, representing the assistant agent's reasoning about user intentions.

\subsection{User emotion thinking}
Both the user agent and the assistant agent are equipped with emotional thinking modules. These modules evaluate the user’s willingness to continue the conversation. This functionality is crucial, as repeatedly inquiring the user can diminish the user experience and potentially discourage further engagement. The emotional thinking of the user agent assesses whether it has been inquired excessively before responding to the assistant agent. If the number of inquiries is too much, the user agent incorporates an emotional expression of reluctance to continue the interaction in its reply, denoted as 
$R^U$, Formally, this can be expressed as:
\begin{equation}
R^U = f(P_e^U,H) 
\end{equation}
where $P_e^U$ represents the prompt input to the LLM $f$ under the current operational context, and $H$ denotes the interaction history.
\par
The emotional thinking module of the assistant agent evaluates whether the user agent's response exhibits reluctance to continue communication. This module outputs the thinking process $E$, defined as:
\begin{equation}
    E = f(P_e^A, R^U)
\end{equation}
If $E$ suggests that the user is unwilling to proceed, the conversation is terminated, and the user's intention is summarized.
\par
Note that in case studies, we explore how to further minimize users' reluctance to continue communicating and improve user experience in practical applications.

\subsection{Reference options thinking}
The reference options thinking module thinks about the most probable intention value that matches the user's expectations for a given intention type $T_i$, based on the current user intention $S_i$. This reasoning process is denoted as $V$, and formally expressed as:
\begin{equation}
    V = f(P_o,S_i,T_i)
\end{equation}
Here, $S_i$ and $T_i$ are the outputs of the user intentions thinking module. The prompt $P_o$ instructs the LLM to analyze the relationship between intention types and values provided by the user and the implicit intention type $T_i$. Subsequently, the LLM is prompted to generate three reference options that are most likely to be accepted by the user. This module allows users to respond by choosing from reference options rather than manually inputting text, thereby improving usability in real-world applications.

\section{Experiment setup}
\subsection{Compared methods}
The data synthesis methods compared in this paper are as follows:
\par
\textbf{Single-Agent:} This method is implemented based on Baize  \cite{xu2023baize}, with prompt optimizations tailored to Chinese travel scenarios. The method guides the LLM to generate dialogues capable of proactively mining implicit intention between the user and the assistant based on the seed data. Additionally, it instructs the LLM to include three reference options for intention values in the assistant's responses. A schematic illustration of the method is provided in Figure \ref{fig:introduction}(a).
\par
\textbf{Dual-Agent:} This method is implemented based on Mistral-interact \cite{qian2024tell}. It employs one LLM as the User Agent and another as the Assistant Agent. The two agents engage in a dialogue based on seed data. The User Agent initiates the interaction by posing the initial inquiry and responding to the Assistant Agent's questions. The Assistant Agent analyzes the user's intention and then requests clarification for implicit intentions, providing three possible intention values as reference options in its response. The schematic diagram of this method is illustrated in Figure \ref{fig:introduction}(b).
\par
The effectiveness of the data synthesis method depends on the ability of the LLM, fine-tuned on the synthetic data, to proactively mine implicit user intention.

\subsection{Training and test sets}
\textbf{Training Set:} A multi-round conversation dataset featuring proactive implicit user intention mining was generated using data synthesis methods, based on the seed data pool described in Section 3.2. The default LLM employed in each synthesis method is Doubao-pro-32K. In SynPT's probabilistic control strategy, the default mean is set to half the number of intentions, with a standard deviation of 2. This dataset serves as the training set for fine-tuning the downstream model. The training dataset includes eight Chinese cities (Beijing, Shanghai, Guangzhou, Chengdu, Xi'an, Hangzhou, Nanning, and Guilin) and incorporates six distinct travel-related tasks: travel planning, attraction recommendation, restaurant recommendation, accommodation booking, shopping venue inquiry, and train ticket reservation.
\par
\textbf{Test set:} We collected seed data from three Chinese cities (Suzhou, Qingdao, and Shenzhen) distinct from those in the training set, then randomly selected 120 samples, with 20 samples per task. Since the evaluation focuses on the proactive implicit user intention mining performance of the assistant model, we created an initial inquiry for each seed data to minimize the influence of user roles. To ensure a fair comparison of data synthesis methods, we deviated from the initial inquiry generation approach described in Section 3.3. Instead, we constructed the user's initial inquiries based on the following rules: 20\% of the initial inquiries contained no intentions, while another 20\% included all intentions.
The remaining 60\% comprised inquiries with a random number of intentions (ranging from 0 to $N$), where $N$ represents the total number of intention types associated with the task.
\par

\subsection{Metrics}
Follow related work \cite{qian2024tell}, the model’s capability for proactively mining implicit user intentions is evaluated using the following metrics. \par
\textbf{Main Metrics}: \par
1.	Vagueness Judgment Accuracy: 
1)	When the user's initial inquiry is clear (i.e., there is no lack of intention), the model is considered correct if it directly summarizes the user's intention. It is deemed incorrect if it instead asks follow-up questions.
2)	When the user's initial inquiry is vague (i.e., there is a lack of intention), the model is considered correct if it asks follow-up questions to obtain the missing details. Conversely, if it directly summarizes the intention without further inquiry, it is marked as incorrect. \par
2.	Implicit Intention Mining Accuracy:
This metric assesses whether the model proactively inquires about all missing intention types in the user's initial inquiry. The model is considered correct only if it successfully mines all such missing types; otherwise, it is marked incorrect. \par
3.	Intention Summarization Accuracy:
This metric evaluates the model's ability to accurately summarize the user's intention types and their corresponding values based on the dialogue history. \par
\textbf{Additional Metrics}: \par
4.	Dialogue Turns:
The total number of interaction turns throughout the dialogue is recorded. This metric evaluates whether the model unnecessarily repeats requests for intentions that the user has already specified.
Dialogue Turns A: This metric counts the total number of interaction turns throughout the entire dialogue.
Dialogue Turns B: This metric counts the number of interaction turns for samples where all three criteria, vagueness judgment, implicit intention extraction, and intention summarization, are correctly met. It is designed to evaluate whether the model redundantly inquires about intentions that have already been explicitly provided by the user. \par
5.	Token Consumption:
This metric measures the total number of tokens consumed by the data synthesis method. \par

\subsection{Protocols}
First, we employ the Low-Rank Adaptation (LoRA) \cite{hulora} method to fine-tune the Qwen2.5-7B-Instruct model using training set data. Fine-tuning is performed using the LLaMA-Factory \cite{zheng2024llamafactory} on A6000 GPUs over 10 epochs. The fine-tuned model serves as the assistant model. Subsequently, we introduce an LLM (default: Doubao-pro-32k) to simulate the user role, which interacts with the assistant model in a conversational manner based on user intention information from the test set. This process generates a dialogue between the simulated user and the assistant. Ultimately, \textbf{we evaluate the dialogue to reflect the assistant model’s capability to proactively mine implicit user intention, thereby determining the effectiveness of the data synthesis method.}  Since implicit intention mining accuracy and intention summarization accuracy cannot be quantified directly, we compute these two metrics using two perspectives: automatic evaluation and human evaluation. The automatic evaluation employs the LLM-as-a-judge methodology, utilizing an LLM to assess the dialogues. For the human evaluation, two independent volunteers were recruited to assess the same conversation recordings based on these two metrics. In cases where their scores diverged, a third evaluator was consulted to determine the final result. To maintain evaluation stability, the temperature parameter for all LLMs was set to 0.01 throughout the experiment.

\section{Research results}

\subsection{Main results}
This subsection adheres to the evaluation protocol outlined in Section 4.4 to evaluate the effectiveness of each data synthesis method. In addition to the methods described in Section 4.1, we evaluate the capacity of LLMs accessed through an API to proactively mine implicit intentions. Specifically, Qwen2.5-7B-Instruct and Doubao-Pro were selected: the former serves as the backbone model, and the latter is the LLM employed by data synthesis methods. Inspired by the success of DeepSeek-R1, we included both DeepSeek-R1 and DeepSeek-V3 \cite{liu2024deepseek} in our comparison. It is important to note that these LLMs, accessed via an API, lack the ability to proactively question. Therefore, we provided prompts and incorporated the task name and intention type of each sample into the context. In contrast, the models trained using data synthesis did not involve any prompts or contextual information. Additionally, we extracted the Assistant Agent from SynPT to include in the comparison because it serves as the distillation target for Qwen-PT. Since the data synthesis method involves a training process, we conducted five independent runs for each experimental setting and report the mean and standard error of the results. For models with prompts, repeated runs were not conducted, as a low temperature coefficient of 0.01 was set to ensure output stability. The experimental results are presented in Table \ref{tab:main}.
{
\setlength{\tabcolsep}{2pt}
\begin{table*}[]
\caption{Performance Comparison of Proactive Mining Implicit User Intentions in Chinese Travel Scenarios. The best results for each category of methods are highlighted in bold. Abbreviations: VJ Acc. = Vagueness Judgment Accuracy; IIM Acc. = Implicit Intention Mining Accuracy; IS Acc. = Intention Summarization Accuracy; Turns A = Dialogue Turns A; Turns B = Dialogue Turns B; Tokens = Token Consumption; (H) = Human Evaluation; Qwen-2.5 = Qwen-2.5-7B-Instruct}
\label{tab:main}
\begin{tabular}{p{2cm}c >{\centering\arraybackslash}p{1.8cm} >{\centering\arraybackslash}p{1.9cm} >{\centering\arraybackslash}p{1.9cm} >{\centering\arraybackslash}p{1.8cm} >{\centering\arraybackslash}p{2.1cm} >{\centering\arraybackslash}p{2cm} >{\centering\arraybackslash}p{2cm}}
\hline
\multicolumn{1}{c}{\multirow{2}{*}{Metrics}} & \multicolumn{5}{c}{w/ Prompt}              & \multicolumn{3}{c}{w/o Prompt}   \\ \cmidrule(l){2-6} \cmidrule(l){7-9} 
\multicolumn{1}{c}{} & Qwen-2.5 & Doubao-Pro & DeepSeek-V3 & DeepSeek-R1 & Assistant Agent (ours) & Single-Agent \cite{xu2023baize} & Dual-Agent \cite{qian2024tell} & SynPT (ours) \\ \hline 
Parameters  & 7B & - & 671B & 671B & - & 7B & 7B & 7B \\ \hline
VJ Acc.            & - & 84.17 & 79.17 & \textbf{90.00} & 89.17 & $67.33\pm0.16$  & $75.83\pm0.59$  & $\mathbf{85.31\pm1.75}$          \\
IIM Acc.           & - & 90.83 & 74.17 & 83.33          & \textbf{99.17} & $88.17\pm1.22$  & $82.00\pm0.97$  & \textbf{$\mathbf{96.83\pm0.72}$} \\
IIM Acc. (H)       & - & 90.83 & 74.17 & 85.83          & \textbf{99.17} & $87.50\pm1.18$  & $82.83\pm0.82$  & $\mathbf{97.17\pm0.28}$ \\
IS Acc.            & - & \textbf{96.67} & 65.83 & 87.50          & 95.83 & $60.17\pm1.61$  & $60.33\pm0.90$  & \textbf{$\mathbf{93.82\pm1.69}$} \\
IS Acc. (H)        & - & 93.33 & 63.33 & 85.83          & \textbf{96.67} & $60.33\pm2.37$  & $62.17\pm0.77$  & $\mathbf{95.50\pm0.28}$ \\ \hline
Turns A            & - & 3.10  & 2.95  & 2.61           & 3.08  & $4.42\pm0.04$   & $3.54\pm0.01$   & $3.07\pm0.05$           \\
Turns B            & - & 3.27  & 2.83  & 2.54           & 3.17  & $5.03\pm0.02$   & $3.48\pm0.10$   & $3.24\pm0.08$           \\
Tokens             & - & -     & -     & -              & -     & 1,470K & 7,710K & 8,017K         \\ \hline
\end{tabular}
\end{table*}

}
\par
The experimental results indicate that SynPT consistently outperforms other data synthesis methods across all evaluated metrics, demonstrating the effectiveness of the proposed strategy. This conclusion is further supported by human evaluation. Moreover, the discrepancy between human and automatic evaluation results is approximately 2 percent, supporting the validity of using an LLM as an evaluator.
In terms of token consumption, although the Single-Agent data synthesis method uses the fewest tokens, it performs the worst in proactive implicit intention mining. This limitation may stem from the simplicity of its synthesis mechanism, which results in poor generalization of the generated training data. Although SynPT consumes only 3.98\% more tokens than the Dual-Agent method, it incorporates additional modules for user emotion thinking and reference option thinking. The overall reduction in token consumption is attributed to the memory stack strategy introduced in the user intention thinking.
From both dialog turn metrics, SynPT yields the lowest values, which indirectly indicates that it reduces the likelihood of repeated inquiries about user-provided intentions. This outcome is attributed to our enhanced initial inquiry generation strategy, which aligns with the true distribution of detail levels.
\par
Next, we analyze the methods incorporating prompts. Despite the assistance of prompts, Qwen-2.5-7B-Instruct fails to proactively mine implicit user intention due to its limited parameter size and weak instruction following capability. However, our data synthesis method enhances the model, transforming it into Qwen-PT, which demonstrates improved performance on the task. 
The Assistant Agent outperforms other methods in implicit intention mining accuracy, while it performs slightly worse than DeepSeek-R1 in vagueness judgment accuracy. For the intention summarization accuracy metric, it performs slightly worse than Doubao-Pro in the automatic evaluation but significantly surpasses Doubao-Pro in the human evaluation. These results demonstrate the strong performance of the Assistant Agent in this task and support the development of a robust Qwen-PT model.
SynPT outperforms prompt-based methods, excluding Assistant Agent, in both implicit intention mining accuracy and intention summarization accuracy (Human). This demonstrates that fine-tuning a smaller LLM with high-quality training data can yield better performance than a larger LLM. However, SynPT demonstrates lower accuracy in vagueness judgment compared to DeepSeek-R1. By analyzing conversation logs, we observed that the user intention thinking module occasionally fails to detect the user's already expressed intention. This oversight results in the erroneous classification of the user's initial inquiry as vague. This limitation stems from the inherent constraints of the LLM employed by SynPT, presenting an area for future optimization.

\subsection{Evaluation of User Emotions}
This subsection evaluates the accuracy of Qwen-PT, a model fine-tuned using SynPT-generated data, in identifying user emotions, particularly in determining whether the user remains willing to continue the interaction. Firstly, the test set for this task was developed. Since the evaluation focuses on the user's emotional response to rhetorical questions, only conversations with two or more rounds were selected from Qwen-PT's dialogue records in Section 4.4. Subsequently, the LLM, Doubao-Pro, is prompted to utilize its inherent knowledge to augment user responses in the dialogue with emotional expressions. The temperature parameter is set to 1 to ensure variability in emotional outputs. Finally, Qwen-PT is evaluated to determine whether it can appropriately terminate further questioning when the user expresses an unwillingness to continue the conversation, while also summarizing the user's stated intention. The experimental results are presented in Table 4.
{
\setlength{\tabcolsep}{2pt}
\begin{table*}
    \caption{Accuracy of Qwen-PT in Judging User Emotions Across Tasks.}
    \label{tab:emotion}
    \centering
    \begin{tabular}{>{\centering\arraybackslash}p{0.07\linewidth}>{\centering\arraybackslash}p{0.1\linewidth}>{\centering\arraybackslash}p{0.14\linewidth}>{\centering\arraybackslash}p{0.14\linewidth}>{\centering\arraybackslash}p{0.14\linewidth}>{\centering\arraybackslash}p{0.14\linewidth}>{\centering\arraybackslash}p{0.14\linewidth}c}
    \toprule
         Task Category & Travel Planning & Attraction Recommendation & Restaurant Recommendation & Accommodation Booking & Shopping Venue Inquiry & Train Ticket Reservation & Average \\ \midrule
        Accuracy & 100.00 & 81.81 & 75.00 & 94.11 & 100.0 & 100.0 & 91.54\\ \bottomrule
    \end{tabular}
\end{table*}
}
\par
The experimental results show that Qwen-PT can basically judge correctly the user's emotion of unwillingness to continue communication with an average accuracy of 91.54\%. In tasks such as travel planning, shopping, and train ticket booking, emotion detection is entirely accurate. However, for other tasks, it does not achieve complete accuracy. We hypothesize that this issue arises from the limited number of emotional training samples in the dataset for attraction recommendation, restaurant recommendation, and accommodation reservation tasks. Since these tasks involve fewer intentions, they generate shorter conversation turns, reducing the likelihood of emotional expressions from the user agent. This limitation can be resolved by increasing the emotional sensitivity of the user agent during data synthesis.

\subsection{Evaluation of Reference Options for Intention Values}
This subsection evaluates the effectiveness of Qwen-PT in providing reference options that align with the user's implicit intentions. In other words, it assesses whether the model can infer reasonable intention values for implicit intention types based on the intention types and intention values already specified by the user. The reference option reasonableness judgment test set is derived from conversation data generated in Section 5.1 and adheres to the following criteria: (1) implicit intention must be present in the initial inquiry (otherwise, the assistant model directly summarizes the user intentions without providing the reference options); (2) the conversation length must be at least two rounds (to validate the consistency between the provided reference option and the user’s subsequent response); and (3) both implicit intention extraction and intention summarization must be confirmed as accurate (to eliminate potential confounding variables in the evaluation). We subsequently follow the LLM-as-a-judge method, employing Doubao-Pro as the evaluation model to assess the reasonableness of the reference option. This assessment is based on a four-level scale, with higher scores indicating a stronger alignment between the reference option and the user's desired intention value:
\begin{itemize}
\item Score 0: The reference option significantly diverges from the user's desired intention.
\item Score 1: The reference option partially aligns with the user's desired intention, but there remains considerable room for improvement.
\item Score 2: The reference option generally matches the user's desired intention, with only minor need for optimization.
\item Score 3: The reference option fully corresponds to the user's desired intention and is both accurate and valuable.
\end{itemize}
\par
Finally, we take the average rating of all samples as an evaluation metric to measure the reasonableness of the reference options for the user intention value. The experimental results are shown in Table \ref{tab:option}.
\begin{table}
    \centering
    \caption{Reasonableness Scores of Reference Options Across Multiple Methods}
    \label{tab:option}
    \begin{tabular}{cccc} \toprule
         & Single-Agent & Dual-Agent & SynPT (ours)\\ \midrule
        Score & 1.14 & 1.61 & 1.71\\ \bottomrule
    \end{tabular}
\end{table}
\par
Experimental results demonstrate that the proposed method surpasses existing data synthesis techniques in terms of reference option rationality. This finding indicates that explicitly thinking reference options enhances the performance of LLMs in this task. However, we observed that the average rating across all methods did not exceed 2. This outcome may be attributed to the substantial variation in user intentions, making it challenging to address a broad spectrum of user needs with only three reference options. While increasing the number of reference options may improve coverage rates, an excessive number of choices could impose selection burdens on users in practice. To address this issue, future research could explore integrating user historical interaction data or profile information to enhance the precision of intention value recommendations.

\section{Discussion}
\subsection{Different number of training samples}
This subsection examines the impact of varying proportions of training data on Qwen-PT's ability to proactively mine implicit user intention. We randomly select different subsets of the training data generated by SynPT and evaluate their performance following the protocol outlined in Section 4.4. The experimental results are presented in Table \ref{tab:proportions}.
{\setlength{\tabcolsep}{2pt}
\begin{table}[]
\caption{The Effect of the Number of Training Samples on Model Performance. The abbreviations are the same as in Table \ref{tab:main}.}
\label{tab:proportions}
\begin{tabular}{@{}lccccccc@{}}
\toprule
\multicolumn{1}{c}{\multirow{2}{*}{Metrics}} & Single-Agent & \multicolumn{1}{l}{Dual-Agent} & \multicolumn{5}{c}{SynPT} \\ \cmidrule(l){2-2}  \cmidrule(l){3-3} \cmidrule(l){4-8}
\multicolumn{1}{c}{}                   & 100\%  & 100\%  & 10\% &20\% & 30\%   & 70\%   & 100\%  \\ \midrule
VJ Acc.            & 67.33  & 75.83  & 72.00 &80.00 & 80.83  & 78.33  & 85.81  \\
IIM Acc. & 88.17  & 82.00  & 78.67 &84.17 & 95.83  & 96.67  & 96.83  \\
IS Acc.      & 60.17  & 60.33  & 49.89 &71.67 & 90.83  & 92.50  & 93.82  \\
Turns A                       & 4.42   & 3.54   & 3.55  &3.40 & 3.29   & 3.13   & 3.07   \\
Turns B                       & 5.03   & 3.48   & 3.45  &3.42 & 3.40   & 3.27   & 3.24   \\
Tokens                      & 1,470K & 7,710K & 763K  &1,550K & 2,325K & 5,465K & 8,017K \\ \bottomrule
\end{tabular}
\end{table}
}
\par
The experimental results indicate that SynPT achieves better performance in proactively mining implicit user intentions as the number of training samples increases. Notably, the model trained with only 30\% of the samples generated by SynPT performs better than models trained with 100\% of the samples generated by Single-Agent or Dual-Agent. This finding further validates the efficacy of our proposed improvement strategy in enhancing the model’s ability to proactively mine implicit user intentions. Moreover, regarding token consumption, only 2,325K tokens are required to generate 30\% of the training samples using SynPT. This amount is lower than that required by Dual-Agent, while SynPT achieves superior performance. Additionally, the token consumption for SynPT to generate 20\% of the training samples is comparable to the amount Single-Agent requires to generate 100\% of the training samples. Nevertheless, models trained with SynPT demonstrate significantly better performance than those trained with Single-Agent in both vagueness judgment accuracy and intention summarization accuracy. These results further underscore the importance of incorporating the memory stack strategy into the user intention thinking module. Regarding the dialogue turns metric, the model trained on only 20\% of the samples generated by SynPT outperformed those trained on 100\% of the samples produced by the Single-Agent or Dual-Agent in reducing repetitive queries for existing explicit intentions. In conclusion, only a small amount of data generated by SynPT is required to significantly enhance the model’s ability to proactively mine implicit user intentions.

\subsection{Different LLMs for User Agent and Assistant Agent}
\begin{figure*}
    \centering
    \includegraphics[width=0.8\linewidth]{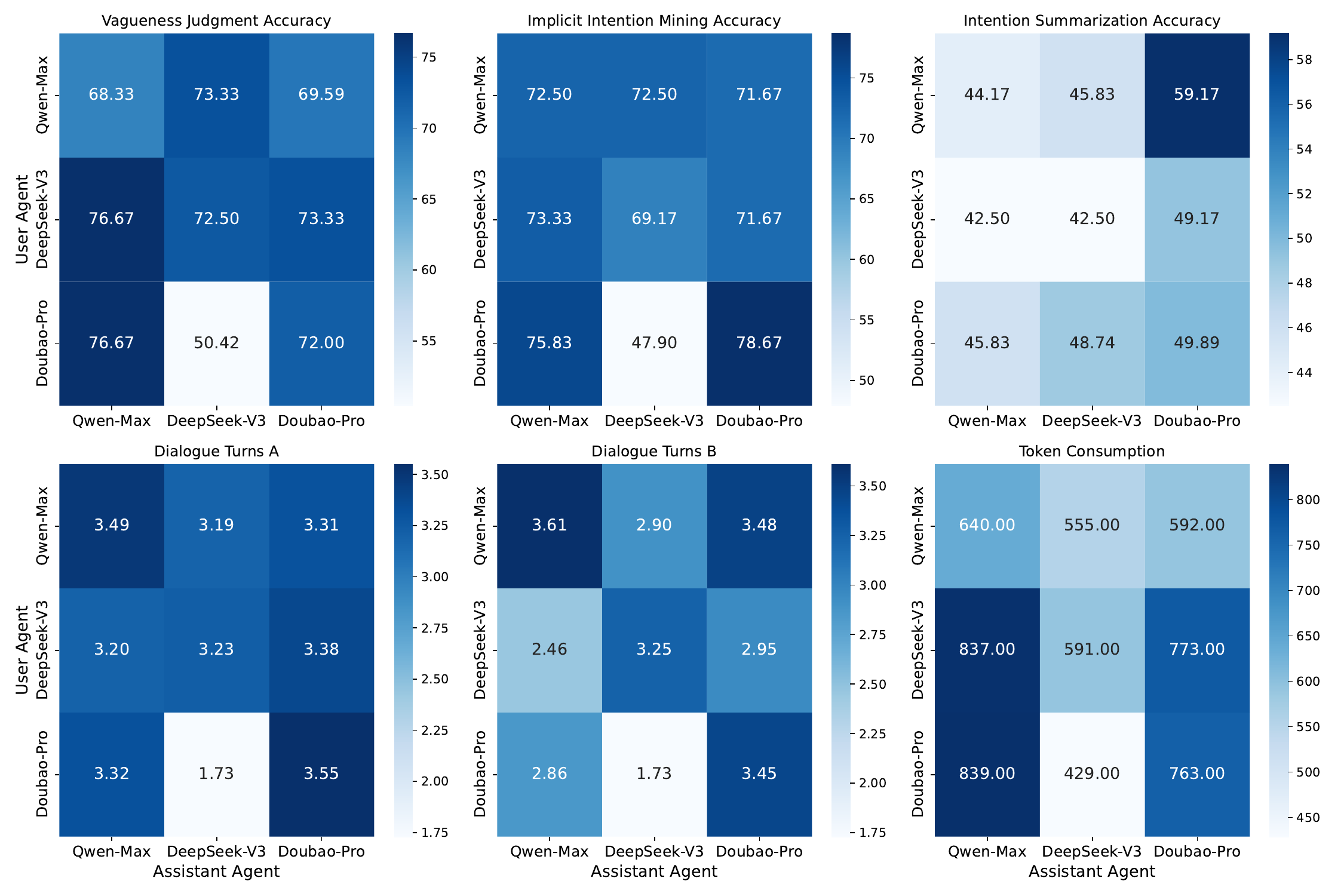}
    \caption{Impact of Using Different LLMs for the User Agent and the Assistant Agent}
    \label{fig:diff_llm}
\end{figure*}

In this subsection, we generate training datasets by configuring different LLMs for the user agent and assistant agent in the SynPT method, with a focus on investigating the impact of LLM combinations on the downstream model's ability to proactively mine implicit user intention. o better highlight the performance differences among LLMs, the experiments specifically employ a 10\% subset of the training data for comparative analysis. Detailed experimental results are presented in Figure \ref{fig:diff_llm}.

\par
The experimental results indicate that when the user agent and the assistant agent use different LLMs, their performance does not exhibit a consistent pattern of change across different metrics. In this study, the default configuration is that both the user agent and the assistant agent use Doubao-Pro, which surpasses other combinations only in implicit intention mining accuracy. These results indicate that the choice of different LLMs affects the model's performance.
\par
To further eliminate the possibility of bias in the evaluation model toward homologous models, specifically whether Doubao-pro would assign higher scores to synthetic data it participated in generating, this study employs DeepSeek-v3 as an alternative evaluation model. The model trained on synthetic data generated by Doubao-pro is reassessed using DeepSeek-v3. Additionally, to maintain evaluation neutrality, user roles are simulated by DeepSeek-v3 during this process. The evaluation results obtained using DeepSeek-v3 show that the downstream model achieves scores of 76.67\%, 90.83\%, and 90.00\% in vagueness judgment accuracy, implicit intention mining accuracy, and intention summarization accuracy, respectively. These scores represent an improvement across all metrics compared to those obtained when Doubao-pro serves as the evaluation model. This result confirms that Doubao-Pro does not exhibit bias toward homologous models during evaluation. Regarding the dialogue turns metric, the combination of the user agent and assistant agent using Doubao-Pro yields a significantly higher value. This suggests that the assistant model trained under this combination may have repeatedly requested intention already provided by the user.
\par
Furthermore, we observed that when the user agent employs Doubao-Pro and the assistant agent utilizes DeepSeek-V3, their performance worsens across main metrics. An analysis of their synthetic data reveals that the initial inquiries generated by Doubao-Pro are relatively brief, which causes DeepSeek-V3 to initiate intention summarization prematurely, even before the user has fully conveyed all implicit intentions. This premature summarization introduces errors into the training data, which subsequently degrades the performance of the downstream model. Notably, this configuration received the lowest score among all combinations on the dialog rounds metric, suggesting that models trained with this pairing significantly reduce the likelihood of repeatedly requesting the user for intentions they have already communicated.
\par
Regarding token consumption, the combination of DeepSeek-V3 as the user agent and Qwen-Max as the assistant agent exhibits the highest usage, yet its performance remains suboptimal. Employing Doubao-Pro for both the user agent and assistant agent results in moderate token consumption. The combination of Doubao-pro as the user agent and DeepSeek-V3 as the assistant agent results in the lowest token consumption. This outcome further indicates that the combination leads to fewer conversational turns and insufficient exchange of intention information.
\par
In summary, the LLM used by the agent in the SynPT method substantially influences the final training outcome. This observation implies that the overall efficacy of SynPT is partially constrained by the inherent capabilities of the LLM and that SynPT is compatible with multiple LLMs. 

\subsection{Ablation experiment}
To evaluate the effectiveness of the probability control mechanism in the initial inquiry generation module and the memory stack mechanism in the user intention reasoning module of SynPT, we conducted ablation experiments. To highlight the differences between the strategies, only 10\% of the training dataset was used. The experimental results are presented in Table \ref{tab:ablation}.

\begin{table}[]
\caption{Results of the ablation experiment}
\label{tab:ablation}
\begin{tabular}{@{}l>{\centering\arraybackslash}p{0.15\linewidth}>{\centering\arraybackslash}p{0.15\linewidth}>{\centering\arraybackslash}p{0.15\linewidth}>{\centering\arraybackslash}p{0.26\linewidth}@{}}
\toprule
Metrics & SynPT & SynPT w/o probability control & SynPT w/o memory stack & SynPT w/o probability control \& memory stack \\ \midrule
VJ Acc.           & 72.00 & 74.58 & 74.14 & 81.51 \\
IIM Acc. & 78.67 & 65.83 & 68.33 & 65.55 \\
IS Acc.       & 49.89 & 46.67 & 40.83 & 35.29 \\
Turns A                       & 3.55  & 2.63  & 3.44  & 2.92  \\
Turns B                       & 3.45  & 2.03  & 2.54  & 1.71  \\
Tokens                      & 763K  & 490K  & 831K  & 652K  \\ \bottomrule
\end{tabular}
\end{table}
\par
The experimental results demonstrate that the model significantly depends on the proposed improvement strategies for both implicit intention mining accuracy and intention summarization accuracy. The removal of either strategy results in a decline in performance. Furthermore, the experimental results demonstrate that removing the memory stack strategy leads to a significant increase in token consumption and a decline in model performance. These findings further confirm the critical role of the memory stack mechanism in compressing contextual information and improving intention mining. Without the probability control strategy, the initial inquiry is more comprehensive, reducing the number of subsequent interaction rounds and resulting in the lowest token consumption. However, this approach negatively impacts the model’s overall performance. Eliminating both the probability control and memory stack strategies leads to moderate token consumption, further supporting the preceding analysis. 
\par
However, regarding vagueness judgment accuracy, the performance of models that omitted the probabilistic control strategy and the memory stack strategy exhibited improvement. Our analysis of the synthetic data indicated that these datasets contained more detailed first-round user consultations and prematurely summarized user intentions before all user inputs were provided. Although this approach improved vagueness judgment accuracy by reducing the likelihood of misclassifying clear intentions as vague, it also diminished its capacity to proactively mine implicit user intentions. This observation is further supported by the lowest value of dialog turns.
For the dialogue turns metric, removing probability control reduces the number of interaction rounds but also decreases performance. In contrast, removing the memory stack has minimal impact on the number of dialogue turns. 
\par
\begin{figure}
    \centering
    \includegraphics[width=0.8\linewidth]{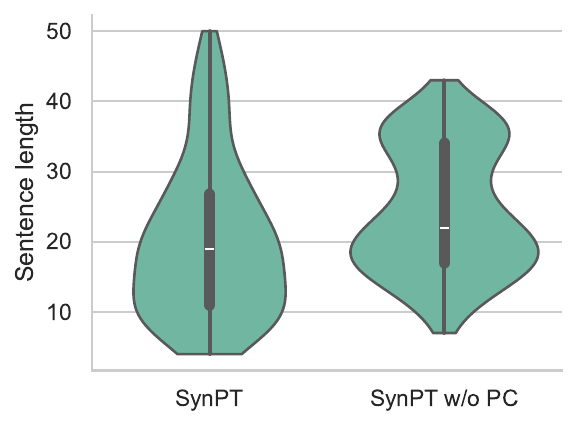}
    \caption{Violin plot of word length distributions in initial inquiries generated by SynPT with and without the probability control (PC) strategy.}
    \label{fig:violin}
\end{figure}

To better understand the impact of the probability control strategy on model behavior, we visualize the word length distribution of the initial inquiries using violin plots, as shown in Figure \ref{fig:violin}. The results indicate that SynPT with probability control produces more diverse query lengths with a lower average length, which aligns with the common user tendency to keep initial queries concise.
In summary, our proposed probabilistic control and memory stack strategies do effectively improve the model's ability to proactively mining implicit user intention.


\subsection{Different backbone networks}
In this experimental, we fine-tune various backbone networks using synthetic data to evaluate their impact on the performance of models in proactively identifying implicit user intention. To better highlight the differences among the backbone networks, only 10\% of the original training dataset was used. We selected small parameter models from the Qwen 2.5-Instruct family and the LLaMA 3.1-8B-Instruct model, which is trained primarily on an English corpus. The results are presented in Table \ref{tab:backbone}.

\begin{table}[]
\caption{Impact of different backbone networks}
\label{tab:backbone}
\begin{tabular}{@{}l>{\centering\arraybackslash}p{0.18\linewidth}>{\centering\arraybackslash}p{0.18\linewidth}>{\centering\arraybackslash}p{0.18\linewidth}>{\centering\arraybackslash}p{0.18\linewidth}@{}}
\toprule
                                       & Qwen 2.5-0.5B-Instruct & Qwen 2.5-3B-Instruct & Qwen 2.5-7B-Instruct & LLaMA 3.1-8B-Instruct \\ \midrule
VJ Acc.            & 68.33                 & 71.67               & 72.00               & 71.67                \\
IIM Acc. & 56.67                 & 72.50               & 78.67               & 86.67                \\
IS Acc.       & 8.33                  & 34.17               & 49.89               & 64.17                \\
Turns A                       & 3.31                  & 3.52                & 3.55                & 3.58                 \\
Turns B                       & 2.00                  & 2.90                & 3.45                & 3.87                 \\ \bottomrule
\end{tabular}
\end{table}

\par
The experimental results indicate that a smaller number of parameters in the backbone network is associated with poorer performance in proactively mining implicit user intentions. Specifically, the Qwen 2.5-0.5B-Instruct model exhibits near-zero accuracy in summarizing user intentions. Additionally, the Qwen 2.5-0.5b-Instruct model exhibits the lowest value on the dialog rounds metric. After reviewing the interaction data, we concluded that this result is not due to the model avoiding the repetition of intentions provided by the user. Instead, it tends to prematurely summarize intentions that are not yet fully expressed.
Furthermore, the Llama 3.1-8B-Instruct \cite{grattafiori2024llama} model outperforms the Qwen 2.5-7B-Instruct model in implicit intention mining accuracy and intention summarization accuracy, suggesting that synthetic datasets from SynPT are compatible with other downstream models.
In conclusion, the benefits of synthetic data vary across different backbone networks.

\subsection{Different training epochs}
In this experiment, we assess the impact of varying training epochs on Qwen-PT's ability to proactively identify implicit user intentions. The results are presented in Table \ref{tab:epoch}.
{
\setlength{\tabcolsep}{7pt}
\begin{table}[]
\caption{Impact of different training epochs}
\label{tab:epoch}
\begin{tabular}{@{}lccccc@{}}
\toprule
                    Metrics       & 2 epochs & 4 epochs & 6 epochs & 8 epochs & 10 epochs \\ \midrule
VJ Acc.            & 71.67    & 78.33    & 81.67    & 82.50    & 85.81     \\
IIM Acc. & 90.00    & 95.83    & 97.50    & 99.17    & 96.83     \\
IS Acc.       & 76.67    & 83.33    & 88.33    & 90.83    & 93.82     \\
Turns A                      & 3.38     & 3.18     & 3.13     & 3.19     & 3.07      \\
Turns B                      & 3.81     & 3.58     & 3.42     & 3.47     & 3.24      \\ \bottomrule
\end{tabular}
\end{table}
}
\par
The experimental results demonstrate that the performance of Qwen-PT in proactively mining implicit user intention improves with increasing training epochs. When trained for eight epochs, the model achieves optimal performance in implicit intention mining accuracy. From the perspective of dialog turn metrics, increasing the number of training epochs helps reduce cases where the model repetitively asks the user about already expressed intentions, thereby improving interaction efficiency and user experience. In summary, the number of training epochs significantly influences the performance of Qwen-PT. Appropriately increasing the number of training epochs can effectively improve the model’s capacity to proactively identify implicit user intentions.

\subsection{Reflection strategies}
Reflective strategies have proven to be highly effective in enhancing the reasoning process of LLMs by facilitating the correction of previously generated errors. In this study, we further investigate the impact of reflective strategies on the ability of downstream models to proactively infer implicit user intentions. Specifically, we implement a reflective strategy following the user intention reasoning module, enabling the LLM to evaluate the accuracy of the user intention reasoning. If it is determined to be inaccurate, the model generates a revised user intention. The experimental results are presented in Table \ref{tab:reflection}.

{
\setlength{\tabcolsep}{3pt}
\begin{table}
    \centering
    \caption{Model performance after adding reflection strategies. The abbreviations are the same as in Table \ref{tab:main}.}
    \label{tab:reflection}
    
    \begin{tabular}{ccccccc}
    \toprule
        Methods & VJ Acc. & IIM Acc. & IS Acc. & Turns A & Turns B & Tokens\\
        \midrule
        SynPT & 72.00 & 78.67 & 49.89 & 3.55 & 3.45 & 763K\\
        Reflection & 73.68 & 67.67 & 35.34 & 3.29 & 2.89 & 1,067K\\
        \bottomrule
    \end{tabular}   
\end{table}
}
\par
The experimental results indicate that introducing the reflection strategy does not enhance the model's ability to proactively mine implicit user intent. Instead, it substantially increases token consumption. An analysis of the training data following the implementation of the reflection strategy reveals that user intention is already accurate in most samples. Consequently, the reasoning process often only involves reflection, omitting the intention correction step. This results in a low proportion of training samples that involve intention correction, making it challenging for the downstream model to learn how to effectively correct inaccurate user intention. Given these findings, SynPT is currently unsuitable for integrating a reflection strategy in this task.

\section{Case Studies}
This section explains the practical application of our proposed SynPT data synthesis method and the Qwen-PT model fine-tuned on synthetic data, with example outputs provided in our GitHub repository.

\subsection{Executing user intentions}
\label{sec:executing user intentions}
The Qwen-PT model is originally designed to summarize user intention rather than execute it. In this subsection, we attempt to enable the model to perform user intention. During the seed data collection process, we gathered both user intentions and their corresponding answers. Specifically, we append the answers to the intention summaries in the dataset, which was generated using our proposed data synthesis method, to create an enhanced training set. This training set is subsequently employed to fine-tune Qwen2.5-7B-Instruct and assess its capability to not only proactively mine the implicit user intention but also effectively resolve their ultimate intention. Sample outputs are available in our GitHub repository. While the model's outputs initially meet expectations, they demonstrate limited effectiveness in resolving user intention. We hypothesize that this limitation arises from the relatively small size of the 7B model, which is not comparable in performance to larger models such as DeepSeek-V3.
\par
To address this problem, we adopt an alternative approach by integrating Qwen-PT with an LLM. Specifically, when Qwen-PT triggers the intention summarization, the resulting summary is used to prompt the LLM to fulfill the user intention. Sample outputs indicate that this approach effectively combines Qwen-PT's ability to proactively infer implicit user intentions with the larger LLM's capability to accurately execute those intentions. Note that experiments employing a larger LLM to proactively mine implicit user intentions have been presented in Section 5.1.

\subsection{An practical agent workflow}
Despite the integration of the user emotion thinking module into SynPT, repeatedly prompting the user may still lead to impatience. To address this issue, we propose an Agent workflow aimed at mitigating the problem. The idea of  this workflow is that the assistant model initially responds to the user's stated intention and subsequently poses a rhetorical question to clarify the user's implicit intention. The workflow is structured as follows.
\par
First, the user's current round query $q_i$ and the previous round response $A_{i-1}^M$ generated by the assistant model are input into Qwen-PT to produce the current round model output $A_{i}^M$. The formal expression is as follows:
\begin{equation}
    A_{i}^M = \text{Qwen-PT}(q_i, A_{i-1}^M)
\end{equation}
\par
Subsequently, other larger LLMs are employed to generate the response $A_i^L$, based on the user current intention. This process is formally represented as follows, where $thought(\cdot)$ denotes the extraction of the user's intention from the assistant's reply.
\begin{equation}
    A_i^L = LLM(thought(A_{i}^M))
\end{equation}
\par
Finally, the response $A_i^L$ generated by the LLM for addressing the user's explicit intention is combined with the implicit intention query extracted from the Qwen-PT response. This process can be formally expressed as follows, where $response(\cdot)$ denotes the function that extracts the implicit intention query from the Qwen-PT response.
\begin{equation}
    A_i = concat(A_i^L, response(A_{i}^M))
\end{equation}
\par
This workflow enables users to receive a response generated according to the provided intention after each query. Additionally, the response includes a question regarding implicit intention at its conclusion.

\subsection{Compatibility of other tasks}
This subsection evaluates the transfer performance of the Qwen-PT model on additional travel-related tasks. Since quantitative evaluation is not feasible in this context, we provide an assessment based on a case study. The model demonstrates limited transferability when applied to new travel-related tasks. It performs well only in scenarios that closely resemble those in the training set, such as airline ticket booking. However, its performance declines markedly in tasks with substantially different distributions, such as purchasing travel insurance or recommending photography guidebooks. An analysis of the dialogue interaction records revealed that the primary cause of the model's inadequate transfer performance is the overfitting of user intention recognition in the first round. This overfitting results in an inaccurate understanding of user intention, which subsequently impairs the performance of later rounds of dialogue generation.
\par
To improve the generalization capability of Qwen-PT, we introduced a broader variety of tourism-related tasks. Since many of these tasks are not explicitly represented on Chinese tourism websites, we generated 14 new categories of seed data using the knowledge capabilities of GPT-4o. Based on this seed data, we employed SynPT to produce a new training set consisting of 700 samples. It is important to note that the SynPT-Dialog training set, discussed in Section 4.2, contains 1500 samples across six task categories.
\par
First, we merged the new dataset with SynPT-Dialog and trained a new model following the protocol described in Section 4.4. Case studies indicate that the model performs well on the 14 newly introduced tourism-related tasks and effectively meets task requirements. Furthermore, we evaluated the model’s performance on general tasks. The results show that it continues to accurately mine implicit user intentions, demonstrating strong adaptability. Notably, the 14 tourism-related task categories contain only 700 samples in total, with approximately 50 samples per category.
\par
We also experimented with using only the first round of dialogue content from the new dataset and merging it with SynPT-Dialog, following the training protocol outlined in Section 4.4, to produce an additional model. Analysis of this model’s outputs on the 14 new tourism-related tasks and generic tasks reveals that it maintains stable performance across all task categories. These findings suggest that the performance degradation of Qwen-PT on transfer tasks primarily results from overfitting during the initial round of user intention recognition. Introducing a limited number of training samples spanning diverse task categories can effectively mitigate this overfitting, thereby enhancing Qwen-PT’s generalization ability across a broader range of task scenarios. Sample outputs are available in our GitHub repository.

\subsection{Compatibility in English-Language Scenarios}
This subsection explores the compatibility performance of the SynPT method in English-language scenarios. Following the experimental design described in Section 7.3, we first constructed an English seed dataset focused on U.S. cities using GPT-4o and translated all relevant SynPT prompts into English. This seed data was then converted into a training set using SynPT, and a new model was trained following the protocol specified in Section 4.4. Sample outputs from this model, which are available in open-source repositories, illustrate its effectiveness in mining implicit user intention in English. These preliminary findings demonstrate the adaptability of the SynPT method to English-language applications.

\section{Conclusion and Future work}
This paper proposes SynPT, a data synthesis method driven by LLMs for proactive mining of implicit user intention in tourism. SynPT is designed to address several limitations observed in existing approaches, including their incompatibility with the Chinese tourism domain, overly concentrated distribution of detail in the initial inquiry, contextual redundancy in the user intention thinking module, and the lack of consideration for user emotion and intention value option reasoning.  
Experimental results demonstrate that SynPT significantly improves downstream models' performance in proactively mining implicit user intention, emotion reasoning, and the rational generation of intention value options. Compared to existing synthetic data generation approaches, SynPT achieves a 9.48\% improvement in vagueness judgment accuracy, an 8.66\% increase in implicit intention mining accuracy, and a remarkable 33.49\% gain in intention summarization accuracy. Furthermore, we conduct a thorough analysis of SynPT's key parameters and offer practical guidelines for parameter selection. Finally, we performed case studies focusing on three critical aspects: reducing user anxiety resulting from repeated questioning, ensuring adaptability to cross-domain tasks, and maintaining compatibility in English-language environments.
\par
Despite its advantages, SynPT still has some limitations. One primary issue is the high token consumption during generation, which may be addressed in future work by developing more efficient memory stack strategies or reusing previously generated components to reduce redundancy. Additionally, models trained on data generated by SynPT may occasionally repeat queries regarding user intentions that have already been provided, which warrants further refinement.

\ifCLASSOPTIONcaptionsoff
\newpage
\fi

\bibliographystyle{IEEEtran}
\bibliography{main}

\end{document}